\documentclass[journal]{IEEEtran}

\usepackage[pdftex]{graphicx}
\usepackage{amsmath}
\usepackage{amssymb}
\usepackage{hyperref}
\hypersetup{
    colorlinks=true,
    linkcolor=black,
    filecolor=black,      
    urlcolor=blue,
}

\usepackage{tabularx}

\begin{document}

\title{SonoNet: Real-Time Detection and Localisation of Fetal Standard Scan Planes in Freehand Ultrasound}

\author{Christian~F.~Baumgartner,~Konstantinos~Kamnitsas,~Jacqueline~Matthew,~Tara~P.~Fletcher,~Sandra Smith,~Lisa~M.~Koch,~Bernhard~Kainz~and~Daniel~Rueckert%
\thanks{This work was supported by the Wellcome Trust IEH Award [102431].}%
\thanks{C.F. Baumgartner, K. Kamnitsas, L.M. Koch, B. Kainz and D. Rueckert were with the Biomedical Image Analysis Group, Department of Computing, Imperial College London, UK.}%
\thanks{S. Smith was with the Division of Imaging Sciences and Biomedical Engineering, King's College London, UK.}%
\thanks{J. Matthew and T.P. Fletcher were with the Division of Imaging Sciences and Biomedical Engineering, King's College London, UK, and the Biomedical Research Centre, Guy's and St Thomas' NHS Foundation, London, UK.}
}

\markboth{Published in IEEE Transactions on Medical Imaging}{} 
\maketitle

\begin{abstract}
Identifying and interpreting fetal standard scan planes during 2D ultrasound mid-pregnancy examinations are highly complex tasks which require years of training. Apart from guiding the probe to the correct location, it can be equally difficult for a non-expert to identify relevant structures within the image. Automatic image processing can provide tools to help experienced as well as inexperienced operators with these tasks. In this paper, we propose a novel method based on convolutional neural networks which can automatically detect 13 fetal standard views in freehand 2D ultrasound data as well as provide a localisation of the fetal structures via a bounding box. An important contribution is that the network learns to localise the target anatomy using weak supervision based on image-level labels only. The network architecture is designed to operate in real-time while providing optimal output for the localisation task. We present results for real-time annotation, retrospective frame retrieval from saved videos, and localisation on a very large and challenging dataset consisting of images and video recordings of full clinical anomaly screenings. We found that the proposed method achieved an average F1-score of 0.798 in a realistic classification experiment modelling real-time detection, and obtained a 90.09\% accuracy for retrospective frame retrieval. Moreover, an accuracy of 77.8\% was achieved on the localisation task. 
\end{abstract}


\IEEEpeerreviewmaketitle

\section{Introduction} 

\IEEEPARstart{A}{bnormal} fetal development is a leading cause of perinatal mortality in both industrialised and developing countries~\cite{salomon2011practice}. Overall early detection rates of fetal abnormalities are still low and are hallmarked by large variations between geographical regions \cite{abuhamad2008automated,binocar2014congenital,hill2015disparities}. 

The primary modality for assessing the fetus' health is 2D ultrasound due to its low cost, wide availability, real-time capabilities and the absence of harmful radiation. However, the diagnostic accuracy is limited due to poor signal to noise ratio and image artefacts such as shadowing. Furthermore, it can be difficult to obtain a clear image of a desired view if the fetal pose is unfavourable. 

Currently, most countries offer at least one routine ultrasound scan at around mid-pregnancy between 18 and 22 weeks of gestation~\cite{salomon2011practice}. Those scans typically involve imaging a number of standard scan planes on which biometric measurements are taken (e.g. head circumference on the trans-ventricular head view) and possible abnormalities are identified (e.g. lesions in the posterior skin edge on the standard sagittal spine view). In the UK, guidelines for selecting and examining these planes are defined in the fetal abnormality screening programme (FASP) handbook \cite{nhs2015FASPHandbook}. 

Guiding the transducer to the correct scan plane through the highly variable anatomy and assessing the often hard-to-interpret ultrasound data are highly sophisticated tasks, requiring years of training~\cite{maraci2014searching}. As a result these tasks have been shown to suffer from low reproducibility and large operator bias~\cite{chan2009volumetric}. Even identifying the relevant structures in a given standard plane image can be a very challenging task for certain views, especially for inexperienced operators or non-experts. At the same time there is also a significant shortage of skilled sonographers, with vacancy rates reported to be as high as 18.1\% in the UK~\cite{sor2014sonographer}. This problem is particularly pronounced in parts of the developing world, where the WHO estimates that many ultrasound scans are carried out by individuals with little or no formal training~\cite{salomon2011practice}.

\begin{figure}[!t]
\centering
\includegraphics[width=1.0\columnwidth]{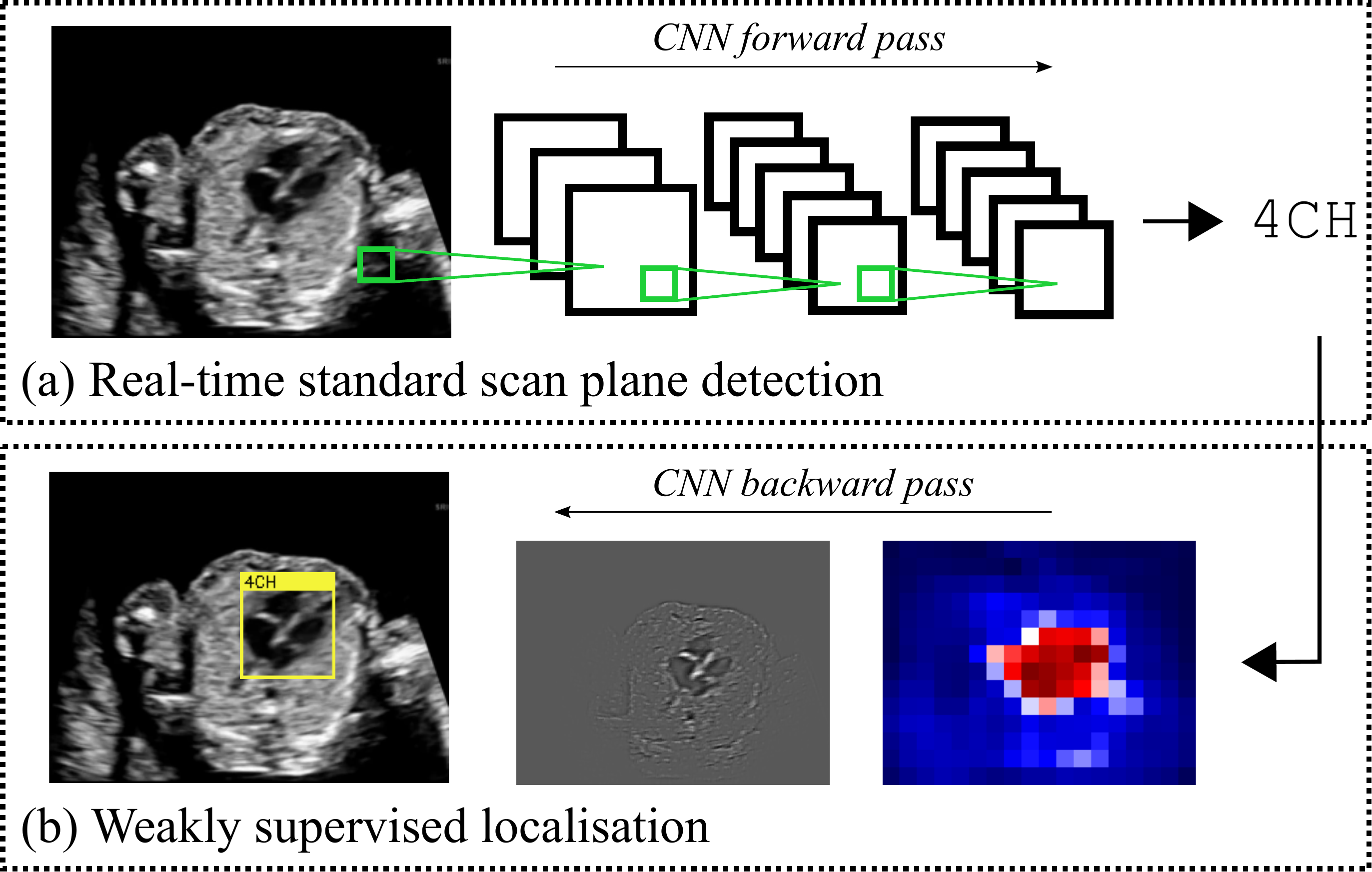}
\caption{Overview of proposed SonoNet: (a) 2D fetal ultrasound data can be processed in real-time through our proposed convolutional neural network to determine if the current frame contains one of 13 fetal standard views (here the 4 chamber view (4CH) is shown); (b) if a standard view was detected, its location can be determined through a backward pass through the network.}
\label{fig:overview}
\end{figure}

\subsection{Contributions} 
 
With this in mind, we propose a novel system based on convolutional neural networks (CNNs) for real-time automated detection of 13 fetal standard scan planes, as well as localisation of the fetal structures associated with each scan plane in the images via bounding boxes. We model all standard views which need to be saved according to the UK FASP guidelines for mid-pregnancy ultrasound examinations, plus the most commonly acquired cardiac views. The localisation is achieved in a weakly supervised fashion, i.e. with only image-level scan plane labels available during training. This is an important aspect of the proposed work as bounding box annotations are not routinely recorded and would be too time-consuming to create for large datasets. Fig.\,\ref{fig:overview} contains an overview of the proposed method. Our approach achieves real-time performance and very high accuracy in the detection task and is the first in the literature to tackle the weakly-supervised localisation task on freehand ultrasound data. All evaluations are performed on video data of full mid-pregnancy examinations. 

The proposed system can be used in a number of ways. It can be employed to provide real-time feedback about the content of a image frame to the operator. This may reduce the number of mistakes made by inexperienced sonographers and could also be applied to automated quality control of acquired images. We also demonstrate how this system can be used to retrospectively retrieve standard views from very long videos, which may open up applications for automated analysis of data acquired by operators with minimal training and make ultrasound more accessible to non-experts. The localisation of target structures in the images has the potential to aid non-experts in the detection and diagnosis tasks. This may be particularly useful for training purposes or for applications in the developing world. Moreover, the saliency maps and bounding box predictions improve the interpretability of the method by visualising the hidden reasoning of the network. That way we hope to build trust into the method and also provide an intuitive way to understand failure cases. Lastly, automated detection and, specifically, localisation of fetal standard views are essential preprocessing steps for other automated image processing such as measurement or segmentation of fetal structures. 

This work was presented in preliminary form in~\cite{baumgartner2016real}. Here, we introduce a novel method for computing category-specific saliency maps, provide a more in-depth description of the proposed methods, and perform a significantly more thorough quantitative and qualitative evaluation of the detection and localisation on a larger dataset. Furthermore, we significantly outperform our results in \cite{baumgartner2016real} by employing a very deep network architecture.  

\subsection{Related work} 

A number of papers have proposed methods to detect fetal anatomy in videos of fetal 2D ultrasound sweeps~(e.g. \cite{maraci2014searching,maraci2015fisher}). In those works the authors have been aiming at detecting the presence of fetal structures such as the skull, heart or abdomen rather than specific standardised scan planes.  

Yaqub et al.~\cite{yaqub2015guided} have proposed a method for the categorisation of fetal mid-pregnancy 2D ultrasound \emph{images} into seven standard scan planes using guided random forests. The authors modelled an ``other'' class consisting of non-modelled standard views. Scan plane categorisation differs significantly from scan plane \emph{detection} since in the former setting it is already known that every image is a standard plane. In standard plane detection on a real-time data stream or video data, standard views must be distinguished from a very large amount of background frames. This is a very challenging task due to the vast amount of possible appearances of the background class. 

Automated fetal standard scan plane detection has been demonstrated for 1--3 standard planes in short videos of 2D fetal ultrasound sweeps~\cite{chen2015automatic,chen2015standard,ni2014standard,ni2013selective}. The earlier of those works rely on extracting Haar-like features from the data and training a classifier such as AdaBoost or random forests on them~\cite{ni2013selective,ni2014standard,zhang2012intelligent}. 

Motivated by advances in computer vision, there has recently been a shift to analyse ultrasound data using CNNs. The most closely related work to ours is that by Chen et al.~\cite{chen2015standard} who employed a classical CNN architecture with five convolutional and two fully-connected layers for the detection of the standard abdominal view. During test time, each frame of the input video was processed by evaluating the classifier multiple times for overlapping image patches. The drawback of this approach is that the classifier needs to be applied numerous times, which precludes the system from running in real-time. In \cite{chen2015automatic}, the same authors extended the above work to three scan planes and a recurrent architecture which took into account temporal information, but did not aim at real-time performance. 

An important distinction between the present study and all of the above works is that the latter used data acquired in single sweeps while we use freehand data. Sweep data are acquired in a fixed protocol by moving the ultrasound probe from the cervix upwards in one continuous motion~\cite{chen2015standard}. However, not all standard views required to determine the fetus' health status are adequately captured using a sweep protocol. For example, imaging the femur or the lips normally requires careful manual scan plane selection. Furthermore, data obtained using the sweep protocol are typically only 2--5 seconds long and consist of fewer than 50 frames~\cite{chen2015standard}. In this work, we consider data acquired during real clinical abnormality screening examinations in a freehand fashion. Freehand scans are acquired without any constraints on the probe motion and the operator moves from view to view in no particular order. As a result such scans can last up to 30 minutes and the data typically consists of over 20,000 individual frames for each case. To our knowledge, automated fetal standard scan plane detection has never been performed in this challenging scenario.

A number of works have been proposed for the supervised localisation of structures in ultrasound. Zhang et al.~\cite{zhang2012intelligent} developed a system for automated detection and fully supervised localisation of the gestational sac in first trimester sweep ultrasound scans. Bridge et al.~\cite{bridge2015object} proposed a method for the localisation of the heart in short videos using rotation invariant features and support vector machines for classification. In more recent work, the same authors have extended the method for the supervised localisation of three cardiac views taking into account the temporal structure of the data~\cite{bridge2017automated}. The method was also able to predict the heart orientation and cardiac phase. To our knowledge, the present work is the first to perform localisation in fetal ultrasound in a weakly supervised fashion.

Although, weakly supervised localisation (WSL) is an active area of research in computer vision (e.g. \cite{ren2016weakly}) we are not aware of any works which attempt to perform WSL in real-time. 

\section{Methods}

\subsection{Data}\label{sec:data} 

Our dataset consisted of 2694 2D ultrasound examinations of volunteers with gestational ages between 18--22 weeks which have been acquired and labelled during routine screenings by a team of 45 expert sonographers according to the guidelines set out in the UK FASP handbook~\cite{nhs2015FASPHandbook}. Those guidelines only define the planes which need to be visualised, but not the sequence in which they should be acquired. The large number of sonographers involved means that the dataset contains a large number of different operator-dependent examination ``styles'' and is therefore a good approximation of the normal variability observed between different sonographers. In order to reflect the distribution of real data, no selection of the cases was made based on normality or abnormality. Eight different ultrasound systems of identical make and model (GE Voluson E8) were used for the acquisitions. For each scan we had access to freeze-frame images saved by the sonographers during the exam. For a majority of cases we also had access to screen capture videos of the entire fetal exam. 

\subsubsection{Image data}

A large fraction of the freeze-frame images corresponded to standard planes and have been manually annotated during the scan allowing us to infer the correct ground-truth (GT) label. Based on those labels we split the image data into 13 standard views. In particular, those included all views required to be saved by the FASP guidelines, the four most commonly acquired cardiac views, and the facial profile view. An overview of the modelled categories is given in Table\,\ref{tab:modelled_categories} and examples of each view are shown in Fig. \ref{fig:detection_collage}.

\begin{table}[t]
   \caption{Overview of the modelled categories.}
 \centering
 \begin{tabular}{ l l }
   \hline
    \multicolumn{2}{l}{\bf Views required by FASP:} \\
    Brain (cb.) & Brain view at the level of the cerebellum \\
    Brain (tv.) & Brain view at posterior horn of the ventricle \\
    Lips    & Coronal view of the lips and nose \\
    Abdominal & Standard abdominal view at stomach level \\
    Kidneys & Axial kidneys view \\
    Femur & Standard femur view \\
    Spine (sag.) & Sagittal spine view \\
    Spine (cor.) & Coronal spine view \\
    \multicolumn{2}{l}{\bf Cardiac views:} \\
    4CH & Four chamber view \\
    3VV & Three vessel view \\
    RVOT & Right ventricular outflow tract \\
    LVOT & Left ventricular outflow tract \\
    \multicolumn{2}{l}{\bf Other:} \\
    Profile & Median facial profile \\
    Background & Non-modelled standard views and background frames \\
   \hline
   \end{tabular}
   \label{tab:modelled_categories}
\end{table}

Additionally, we modelled an ``other'' class using a number of views which do not need to be saved according to the FASP guidelines but are nevertheless often recorded at our partner hospital. Specifically, the ``other'' class was made up from the arms, hands and feet views, the bladder view, the diaphragm view, the coronal face view, the axial orbits view, and views of the cord-insert, cervix and placenta. Overall, our dataset contained 27731 images of standard views and 6856 of ``other'' views. The number of examples for each class ranged from 543 for the profile view to 4868 for the brain (tv.) view. Note that a number of the cases were missing some of the standard planes while others had multiple instances of the same view acquired at different times.

\subsubsection{Video data}

In addition to the still images, our dataset contained 2638 video recordings of entire fetal exams, which were on average over 13 minutes long and contained over 20000 frames. 2438 of those videos corresponded to cases for which image data was also available. Even though in some examinations not all standard views were manually annotated, we found that normally all standard views did appear in the video.

It was possible to find each freeze-frame image in its corresponding video if the latter existed. As will be described in more detail in Sec. \ref{sec:training} we used this fact to augment our training dataset in order to bridge the small domain gap between image and video data. Specifically, the corresponding frames could be found by iterating through the video frames and calculating the image distance of each frame to the freeze-frame image. The matching frame was the one with the minimum distance to the freeze-frame. 

As is discussed in detail in Sec.  \ref{sec:experiments}, all evaluations were performed on the video data in order to test the method in a realistic scenario containing motion and a large number of irrelevant background frames.

\subsection{Preprocessing}\label{sec:preprocessing}

\begin{table}[t]
   \caption{Data preprocessing summary.}
 \centering
 \begin{tabularx}{\columnwidth}{ l l }
   \hline
   {\bf Preprocessing step:} & {\bf Target data:} \\
   \hline
    1) Remove Doppler and split views & Images \& videos \\
    2) Sample random background frames & Videos \\
    3) Crop field of view and rescale & Images \& sampled frames \\
    4) Inpaint labels and annotations & Images \& sampled frames \\
    5) Split into training and test sets & All data \\

   \hline
   \end{tabularx}
   \label{tab:preprocessing}
\end{table}

The image and video data were preprocessed in five steps which are summarised in Table\,\ref{tab:preprocessing} and will be discussed in detail in the following.

Since, in this study, we were only interested in structural images we removed all freeze-frame images and video frames containing colour Doppler overlays from the data. We also removed video frames and images which contained split views showing multiple locations in the fetus simultaneously.

To prevent our algorithm from learning the manual annotations placed on the images by the sonographers rather than from the images themselves, we removed all the annotations using the inpainting algorithm proposed in~\cite{telea2004image}.

We rescaled all image and frame data and cropped a 224x288 region containing most of the field of view but excluding the vendor logo and ultrasound control indicators. We also normalised each image by subtracting the mean intensity value and dividing by the image pixel standard deviation. 

In order to tackle the challenging scan plane detection scenario in which most of the frames do not show any of the standard scan planes, a large set of background images needed to be created. The data from the ``other'' classes mentioned above were not enough to model this highly varied category. 

Note that our video data contained very few frames showing standard views and the majority of frames were background. Thus, it was possible to create the background class by randomly sampling frames from the available video recordings. Specifically, we sampled 50 frames from all training videos and 200 frames from all testing videos. While we found that 50 frames per case sufficed to capture the full variability of the background class during training, we opted for a larger number of background frames for the test set in order to evaluate the method in a more challenging and realistic scenario. This resulted in a very large background class with 110638 training images and 105611 testing images. Note that operators usually hold the probe relatively still around standard planes, while the motion is larger when they are searching for views. Thus, in order to decrease the chance of randomly sampling actual standard planes, frames were only sampled where the probe motion, i.e. image distance to previous video frame, was above a small threshold. Note that the location in the video of some of the standard scan planes could be determined by comparing image distances to the freeze frames as described earlier (see Sec.~\ref{sec:data}). However, this knowledge could not be used to exclude all standard views for the background class sampling because it only accounted for a very small fraction of standard views in the video. The videos typically contained a large number of unannotated standard views in the frames before and after the freeze frame, and also in entirely different positions in the video.

The images from the ``other'' category were also added to the background class. Overall the dataset including the background class had a substantial (and intentional) class imbalance between standard views and background views. For the test set the standard view to background ratios were between 1:138 and 1:1148, depending on the category.

In the last step, we split all of the cases into a training set containing 80\% of the cases and test set containing the remaining 20\%. The split was made on the case level rather than the image level to guarantee that no video frames originating from test videos were used for training. Note that not all cases contained all of the standard views and as a result the ratios between test and training images were not exactly 20\% for each class.

\subsection{Network architecture}

\begin{figure*}[!t]
\centering
\includegraphics[width=1.0\textwidth]{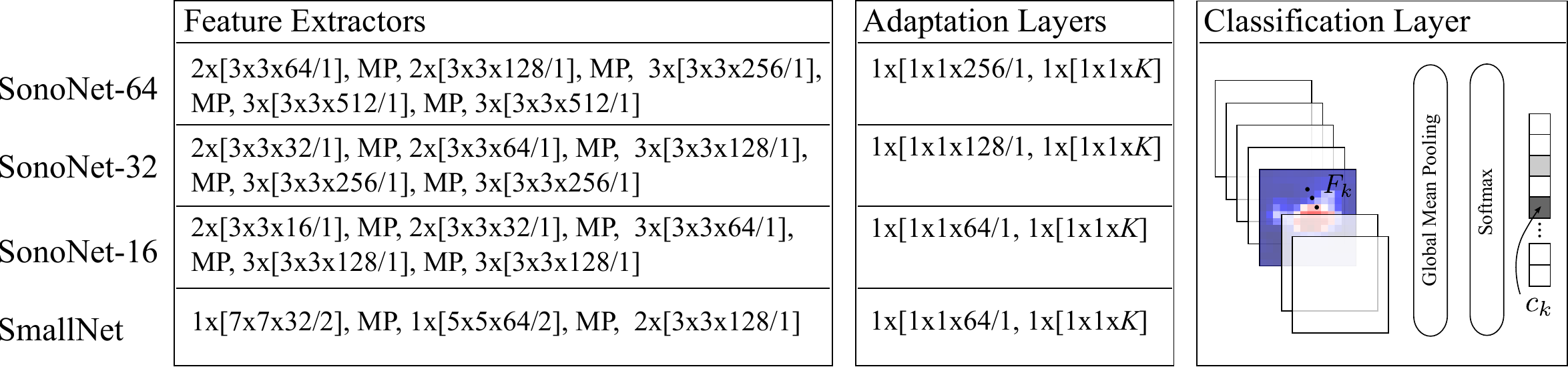}
\caption{Overview of proposed network architectures. Each network consists of a feature extractor, an adaptation layer, and the final classification layer. All convolutional operations are denoted by squared brackets. Specifically, we use the following notation: [\emph{kernelsize} x \emph{number of kernels} / \emph{stride}]. The factor in front of the squared brackets indicates how many times this operation is repeated. Max-pooling is always performed with a kernel size of 2x2 and a stride of 2 and is denoted by MP. All convolutions are followed by a batch normalisation layer before the ReLu activation, except the SmallNet network, for which no batch normalisation was used.}  
\label{fig:architectures}
\end{figure*}    

Our proposed network architecture, the sonography network or SonoNet, is inspired by the VGG16 model which consists of 13 convolutional layers and 3 fully-connected layers~\cite{simonyan2014very}. However, we introduce a number of key changes to optimise it for the real-time detection and localisation tasks. The network architectures explored in this work are summarised in Fig. \ref{fig:architectures}.

Generally, the use of fully-connected layers restricts the model to fixed image sizes which must be decided during training. In order to obtain predictions for larger, rectangular input images during test time, typically the network is evaluated multiple times for overlapping patches of the training image size. This approach was used, for example, in some related fetal scan plane detection works~\cite{chen2015automatic,chen2015standard}. Fully convolutional networks, in which the fully-connected layers have been replaced by convolutions, can be used to calculate the output to arbitrary images sizes much more efficiently in a single forward pass. The output of such a network is no longer a single value for each class, but rather a class score map with a size dependent on the input image size~\cite{lin2013network}. In order to obtain a fixed-size vector of class scores, the class score maps can then be spatially aggregated using the sum, mean or max function to obtain a single prediction per class. Fully convolution networks have been explored in a number of works in computer vision (e.g. \cite{oquab2015object,kang2014fully}), and in medical image analysis, for example for mitosis detection \cite{chen2016mitosis}.

Simonyan et al.~\cite{simonyan2014very} proposed training a traditional model with fully-connected layers, but then converting it into a fully convolutional architecture for efficient testing. This was achieved by converting the first fully-connected layer to a convolution over the full size of the last class score map (i.e. a 7x7 convolution for the VGG16 network), and the subsequent ones to 1x1 convolutions. In the case of 224x288 test images this would produce 1x14 class score maps for each category. 

In this work we use the spatial correspondence between class score maps with the input image to obtain localised category-specific saliency maps (see Sec. \ref{sec:localisation}). Consequently, it is desirable to design the network such that it produces class score maps with a higher spatial resolution. To this end, we forgo the final max-pooling step in the VGG16 architecture and replace all the fully-connected layers with two 1x1 convolution layers. Following the terminology introduced by Oquab et al.~\cite{oquab2015object}, we will refer to those 1x1 convolutions as adaptation layers. The output of those layers are $K$ class score maps $F_k$, where $K$ is the number of modelled classes (here $K=14$, i.e. 13 standard views plus background). We then aggregate them using the mean function to obtain a prediction vector which is fed into the final softmax. In this architecture the class score maps $F_k$ have a size of 14x18 for an 224x288 input image. Note that each neuron in $F_k$ corresponds to a receptive field in the original image creating the desired spatial correspondence with the input image. During training, each of the neurons learns to respond to category-specific features in its receptive field. Note that the resolution of the class score maps is not sufficient for accurate localisation. In Sec. \ref{sec:localisation} we will show how $F_k$ can be upsampled to the original image resolution using a backpropagation step to create category-specific saliency maps.

The design of the last two layers of the SonoNet is similar to work by Oquab et al.~\cite{oquab2015object}. However, in contrast to that work, we aggregate the final class score maps using the mean function rather than the max function. Using the mean function incorporates the entire image context for the classification while using the max function only considers the receptive field of the maximally activated neuron. While max pooling aggregation may be beneficial for the localisation task~\cite{pinheiro2015image,oquab2015object}, we found the classification accuracy to be substantially lower using that strategy. 

Since we are interested in operating the network in real-time, we explore the effects of reducing the complexity of the network on inference times and detection accuracy. In particular, we investigate three versions of the SonoNet. The SonoNet-64 uses the same architecture for the first 13 layers as the VGG16 model, with 64 kernels in the first convolutional layer. We also evaluate the SonoNet-32 and the SonoNet-16 architectures, where the number of all kernels in the network is halved and quartered, respectively.

In contrast to the VGG16 architecture, we include batch normalisation in every convolutional layer~\cite{ioffe2015batch}. This allows for much faster training because larger learning rates can be used. Moreover, we found that for all examined networks using batch normalisation produced substantially better results. 

In addition to the three versions of the SonoNet, we also compare to a simpler network architecture which is loosely inspired by the AlexNet \cite{krizhevsky2012imagenet}, but has much fewer parameters. This is also the network which we used for our initial results presented in \cite{baumgartner2016real}. Due to the relatively low complexity of this network compared to the SonoNet, we refer to it as SmallNet. 

\subsection{Training}\label{sec:training} 

We trained all networks using mini-batch gradient descent with a Nesterov momentum of 0.9, a categorical cross-entropy loss and with an initial learning rate of $0.1$. We subsequently divided the learning rate by 10 every time the validation error stopped decreasing. In some cases we found that a learning rate of $0.1$ was initially too aggressive to converge immediately. Therefore, we used a warm-up learning rate of $0.01$ for 500 iterations~\cite{he2015deep}. Since the SmallNet network did not have any batch normalisation layers it had to be trained with a lower initial learning rate of $0.001$. % 

Note that there is a small domain gap between the annotated image data and the video data we use for our real-time detection and retrospective retrieval evaluations. Specifically, the video frames are slightly lower resolution and have been compressed. In order to overcome this, we automatically identified all frames from the training videos which corresponded to the freeze-frame images in our training data. However, as mentioned in Sec. \ref{sec:data} not all cases had a corresponding video, such that the frame dataset consisted of fewer instances than the image dataset. To make the most of our data while ensuring that the domain gap is bridged, we combined all of the images \emph{and} the corresponding video frames for training. We used 20\% of this combined training dataset for validation. 

In order to reduce overfitting and make the network more robust to varying object sizes we used scale augmentation~\cite{simonyan2014very}. That is, we extracted square patches of the input images for training by randomly sampling the size of the patch (between 174x174 and 224x224) and then scaling it up to 224x224 pixels. To further augment the dataset, we randomly flipped the patches in the left-right direction, and rotated them with a random angle between $-25^\circ$ and $25^\circ$. 

The training procedure needed to account for the significant class imbalance introduced by the randomly sampled background frames. Class imbalance can be addressed either by introducing an asymmetric cost-function, by post-processing the classifier output, or by sampling techniques \cite{zhou2006training,he2009learning}. We opted for the latter approach which can be neatly integrated with mini-batch gradient descent. We found that the strategy which produced the best results was randomly sampling mini-batches that were made up of the same number of standard planes and background images. Specifically, we used 2 images of each of the 13 standard planes and 26 background images per batch. 

The optimisation typically converged after around 2 days of training on a Nvidia GeForce GTX 1080 GPU. 

\subsection{Frame annotation and retrospective retrieval}

After training we fed the network with cropped video frames with a size of 224x288. This resulted in $K$ class score maps $F_k$ with a size of 14x18. Those where averaged in the mean pooling layer to obtain a single class score $a_k$ for each category $k$. The softmax layer then produced the class confidence $c_k$ of each frame. The final prediction was given by the output with the highest confidence. 

For retrospective frame retrieval we calculated and recorded the confidence $c_k$ for each class over the entire duration of an input video. Subsequently, we retrieved the frame with the highest confidence for each class. 

\subsection{Weakly supervised localisation}\label{sec:localisation}

After determining the 14x18 class score maps $F_k$ and the image category in a forward pass through the network, the fetal anatomical structures corresponding to that category can then be localised in the image. A coarse localisation could already be achieved by directly relating each of the neurons in $F_k$ to its receptive field in the original image. However, it is also possible to obtain pixel-wise maps containing information about the location of class-specific target structures at the resolution of the original input images. This can be achieved by calculating how much each pixel influences the activation of the neurons in $F_k$. Such maps can be used to obtained much more accurate localisation. Examples of $F_k$ and corresponding saliency maps are shown in Fig. \ref{fig:saliency_full_vs_part}.

In the following we will show how category-specific saliency and confidence maps can be obtained through an additional backward pass through the network. Secondly, we show how to post-process the saliency maps to obtain confidence maps from which we then extract a bounding box around the detected structure.

\subsubsection{Category-specific saliency maps} 

Generally, category-specific saliency maps $S_k$ can be obtained by computing how much each pixel in the input image $X$ influences the current prediction. This is equivalent to calculating the gradient of the last activation before the softmax $a_k$ with respect to the pixels of the input image $X$. 

\begin{equation}\label{eq:sal0}
S_k = \frac{\partial a_k}{\partial X} 
\end{equation}

\begin{figure}[!t]
\centering
\includegraphics[width=1.0\columnwidth]{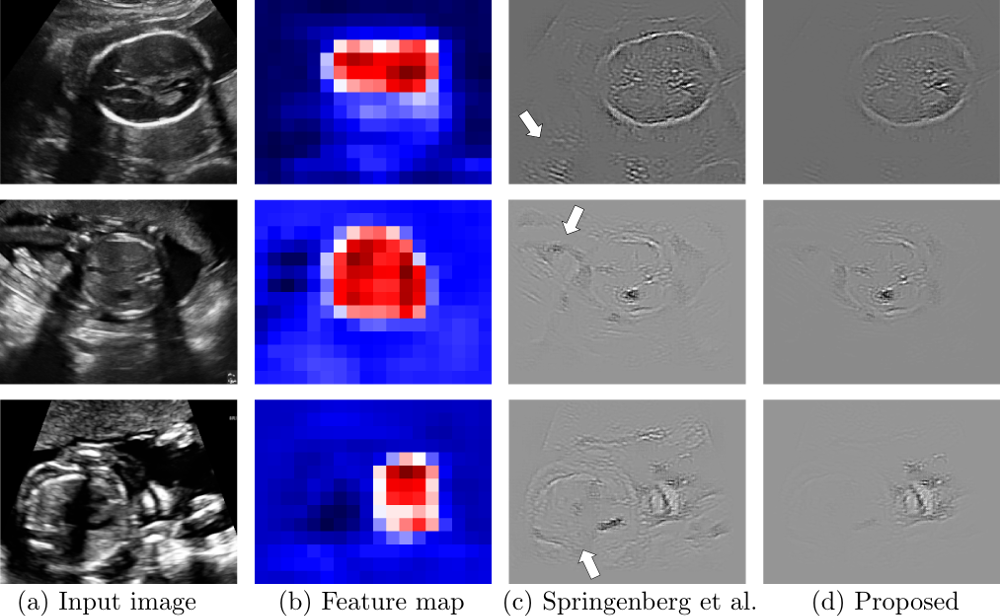}
\caption{Examples of saliency maps. Column (a) shows three different input frames, (b) shows the corresponding class score maps $F_k$ obtained in the forward pass of the network, (c) shows saliency maps obtained using the method by Springenberg et al.~\cite{springenberg2014striving} and (d) shows the saliency maps resulting from our proposed method. Some of the unwanted saliency artefacts are highlighted with arrows in (c). }
\label{fig:saliency_full_vs_part}
\end{figure}

The gradient can be obtained efficiently using a backward pass through the network \cite{simonyan2013deep}. Springenberg et al.~\cite{springenberg2014striving} proposed a method for performing this back-propagation in a \emph{guided} manner by allowing only error signals which contribute to an increase of the activations in the higher layers (i.e. layers closer to the network output) to back-propagate. In particular, the error is only back-propagated through each neuron's ReLU unit if the input to the neuron $x$, as well as the error in the next higher layer $\delta_n$ are positive. That is, the back-propagated error $\delta_{n-1}$ of each neuron is given by 
\begin{equation}
\delta_{n-1}=\delta_n\sigma(x)\sigma(\delta_n),
\end{equation}
where $\sigma(\cdot)$ is the unit step function. Examples of saliency maps obtained using this method are shown in Fig. \ref{fig:saliency_full_vs_part}b. It can be observed that those saliency maps, while highlighting the fetal anatomy, also tend to highlight background features, which adversely affects automated localisation. 

In this work, we propose a method to generate significantly less noisy, localised saliency maps by taking advantage of the spatial encoding in the class score maps $F_k$. As can be seen in Fig. \ref{fig:saliency_full_vs_part}a, the class score maps can be interpreted as a coarse confidence map of the object's location in the input frame. In particular, each neuron $h_k^{n}(X)$ in $F_k$ has a receptive field in the original image $X$. In our preliminary work \cite{baumgartner2016real}, we backpropagated the error only from a fixed percentile $P$ of the most highly activated neurons in $F_k$ to achieve a localisation effect. However, this required heuristic selection of $P$. In this paper, we propose a more principled approach. 

Note that very high or very low values in the saliency map mean that a change in that pixel will have a large effect on the classification score. However, those values do not necessarily correspond to high activations in the class score map. For example, an infinitesimal change in the input image may not have a very large impact if the corresponding output neuron is already very highly activated. Conversely, another infinitesimal change in the input image may have a big impact on a neuron with low activation, for example by making the image look less like a competing category. To counteract this, we preselect the areas of the images which are likely to contain the object based on the class score maps and give them more influence in the saliency map computation. More specifically, we use the activations $h_k^{n}(X)$ in $F_k$ to calculate the saliency maps as a weighted linear combination of the influence of each of the receptive fields of the neurons in $F_k$. In this manner, regions corresponding to highly activated neurons will have more importance than neurons with low activations in the resulting saliency map. In the following, we drop the subscripts for the category $k$ for conciseness. We calculate the saliency map $S$ as 
\begin{equation}\label{eq:sal1}
S = \sum_n h_{>0}^n(X) \frac{\partial h^n(X)}{\partial X}, 
\end{equation}
where $h_{>0}^n$ are the class score map activations thresholded at zero, i.e. $h_{>0}^n = h^n\sigma(h^n)$. By thresholding at zero we essentially prevent negative activations from contributing to the saliency maps. Note that it is not necessary to back-propagate for each neuron $h^n$ separately. In fact, the saliency can still be calculated in a single back-propagation step, which can be seen by rewriting Eq. \ref{eq:sal1} as
\begin{equation}\label{eq:sal2}
S = \sum_n \frac{1}{2}\frac{\partial (h_{>0}^n(X))^2}{\partial X} = \frac{1}{2}\frac{\partial\mathbf{e}^T F_{>0} \circ F_{>0} \mathbf{e}}{\partial X},
\end{equation}
where $F_{>0}$ is the class score map thresholded at zero, $\circ$ is the element-wise matrix multiplication and $\mathbf{e}$ is a vector with all ones. The first equality stems from the chain-rule and the observation that $h_{>0}^n h^n=h_{>0}^n h_{>0}^n$, and the second equality stems from rewriting the sum in matrix form. 

Examples of saliency maps obtained using this method are shown in Fig. \ref{fig:saliency_full_vs_part}c. It can be seen that the resulting saliency maps are significantly less noisy and the fetal structures are easier to localise compared to the images obtained using the approach presented in \cite{springenberg2014striving}.

\subsubsection{Bounding box extraction}  

\begin{figure}[!t]
\centering
\includegraphics[width=1.0\columnwidth]{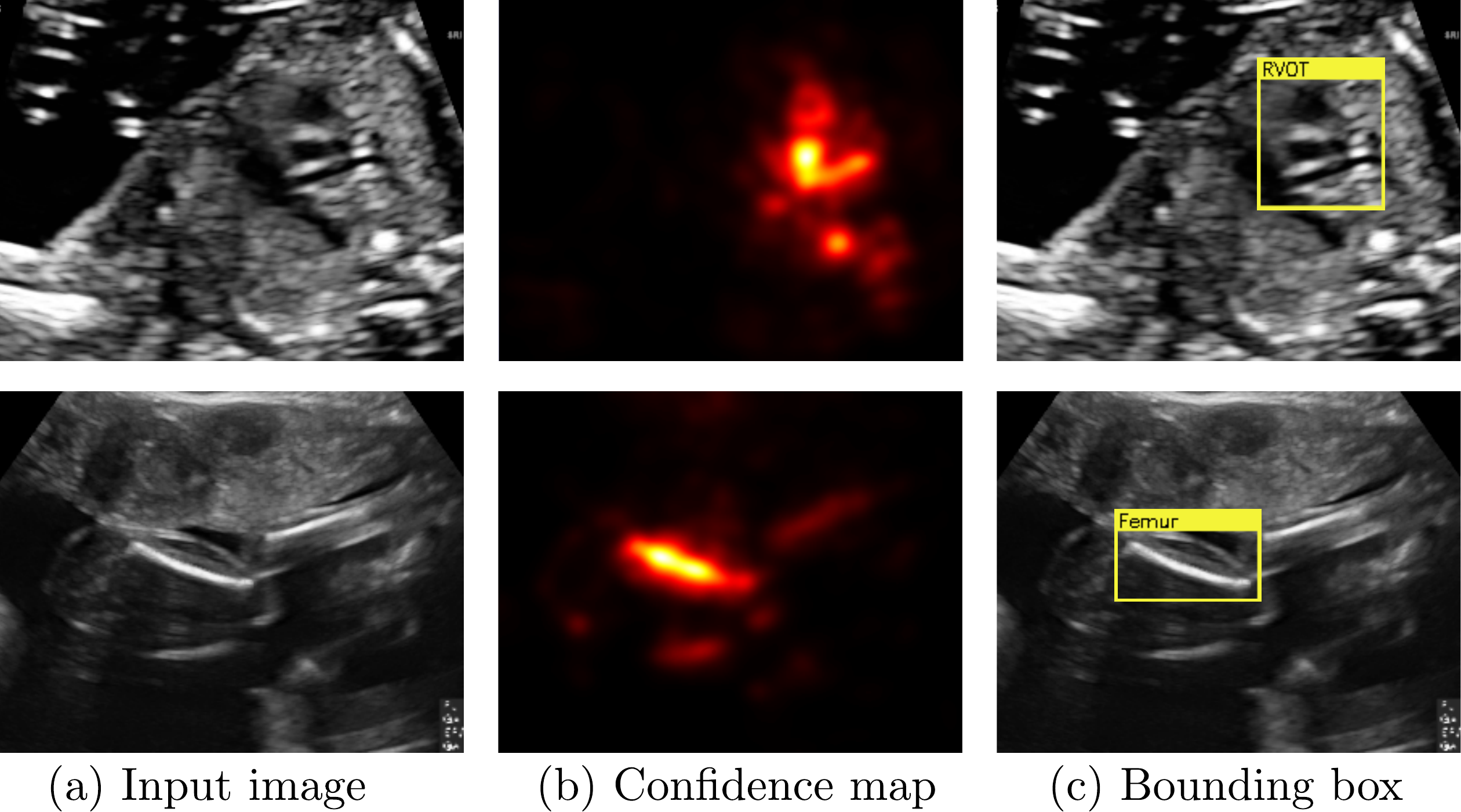}
\caption{Examples of saliency map post-processing for two challenging views: (a) shows two input images, (b) shows the resulting confidence maps for those images, and (c) shows the resulting bounding boxes.}
\label{fig:saliency_post_processing}
\end{figure}

Next, we post-process saliency maps obtained using Eq. \ref{eq:sal2} to obtain confidence maps from which we then calculate bounding boxes. In a first step, we take the absolute value of the saliency map $S$ and blur it using a $5x5$ Gaussian kernel. This produces confidence maps of the location of the structure in the image such as the ones shown in Fig. \ref{fig:saliency_post_processing}b. Note that even though both structures are challenging to detect on those views, the confidence maps localise them very well, despite artefacts (shadows in row 1) and similar looking structures (arm in row 2).

Due to the way the gradient is calculated structures that appear dark in the images (such as cardiac vessels) will usually have negative saliencies and structures that appear bright (bones) will usually have positive saliencies in $S_k$. We exploit this fact to introduce some domain knowledge into the localisation procedure. In particular, we only consider positive saliencies for the femur, spine and lips, and we only consider negative saliencies for all cardiac views. We use both positive and negative for the remainder of the classes. 

Next, we threshold the confidence maps using the Isodata thresholding method proposed in~\cite{el2010images}. In the last step, we take the largest connected component of the resulting mask and fit the minimum rectangular bounding box around it. Two examples are shown in Fig. \ref{fig:saliency_post_processing}c.  

\section{Experiments and Results}\label{sec:experiments}

\subsection{Real-time scan plane detection}

In order to quantitatively assess the detection performance of the different architectures we evaluated the proposed networks on the video frame data corresponding to the freeze-frames from the test cohort including the large amount of randomly sampled background frames. We measured the algorithm's performance using the precision (TP / (TP + FP)) and recall (TP / (TP + FN)) rates as well as the F1-score, which is defined as the harmonic mean of the precision and recall. In Table\,\ref{tab:detection_scores_all} we report the average scores for all examined networks. Importantly, the average was not weighted by the number of samples in each category. Otherwise, the average scores would be dominated by the massive background class. 

In Table\,\ref{tab:frame_rates} we furthermore report the frame rates achieved on a Nvidia Geforce GTX 1080 GPU\footnote{The system was furthermore comprised of an Intel Xeon CPU E5-1630 v3 at 3.70GHz and 2133 MHz DDR4 RAM.} for the detection task alone, the localisation task alone and both of them combined. There is no consensus in literature over the minimum frame rate required to qualify as real-time, however, a commonly used figure is 25 frames per second (fps), which coincides with the frame rate our videos were recorded at.  

\begin{table}[t]
   \caption{Classification scores for the four examined network architectures.}
 \centering
 \begin{tabular}{ l c c c c }
   \hline
   { Network} & { Precision} & { Recall} & { F1-score}\\
   \hline
    SonoNet-64 & {\bf 0.806} & 0.860 & {\bf 0.828} \\
    SonoNet-32 & 0.772 & 0.843 & 0.798 \\
    SonoNet-16 & 0.619 & {\bf 0.900} & 0.720 \\
    SmallNet   & 0.354 & 0.864 & 0.461 \\
   \hline
   \end{tabular}
   \label{tab:detection_scores_all}
\end{table}

 \begin{table}[t]
   \caption{Frame rates in fps for the detection (forward pass), localisation (backward pass) and the two combined.}
 \centering
 \begin{tabular}{ l c c c c }
   \hline
   { Network} & { Detection } & { Localisation } & { Det. \& Loc.} \\
   \hline
    SonoNet-64 & 70.4 & 21.9 & 16.7 \\
    SonoNet-32 & 125.4 & 35.8 & 27.9 \\
    SonoNet-16 & 196.7 & 55.9 & 43.5 \\
    SmallNet   & 574.1 & 226.0 & 162.2 \\
   \hline
   \end{tabular}
   \label{tab:frame_rates}
\end{table}

From Tables\,\ref{tab:detection_scores_all} and \ref{tab:frame_rates} it can be seen that SonoNet-64 and SonoNet-32 performed very similarly on the detection task with SonoNet-64 obtaining slightly better F1-scores, but failing to perform the localisation task at more than 25 fps. The SonoNet-32 obtained classification scores very close to the SonoNet-64 but at a substantially lower computational cost, achieving real-time in both the detection and localisation tasks. Further reducing the complexity of the network led to more significant deteriorations in detection accuracy as can be seen from the SonoNet-16 and the SmallNet network. Thus, we conclude that the SonoNet-32 performs the best out of the examined architectures which achieve real-time performance and we use that architecture for all further experiments and results.

\begin{figure}
\centering
\includegraphics[width=1.0\columnwidth]{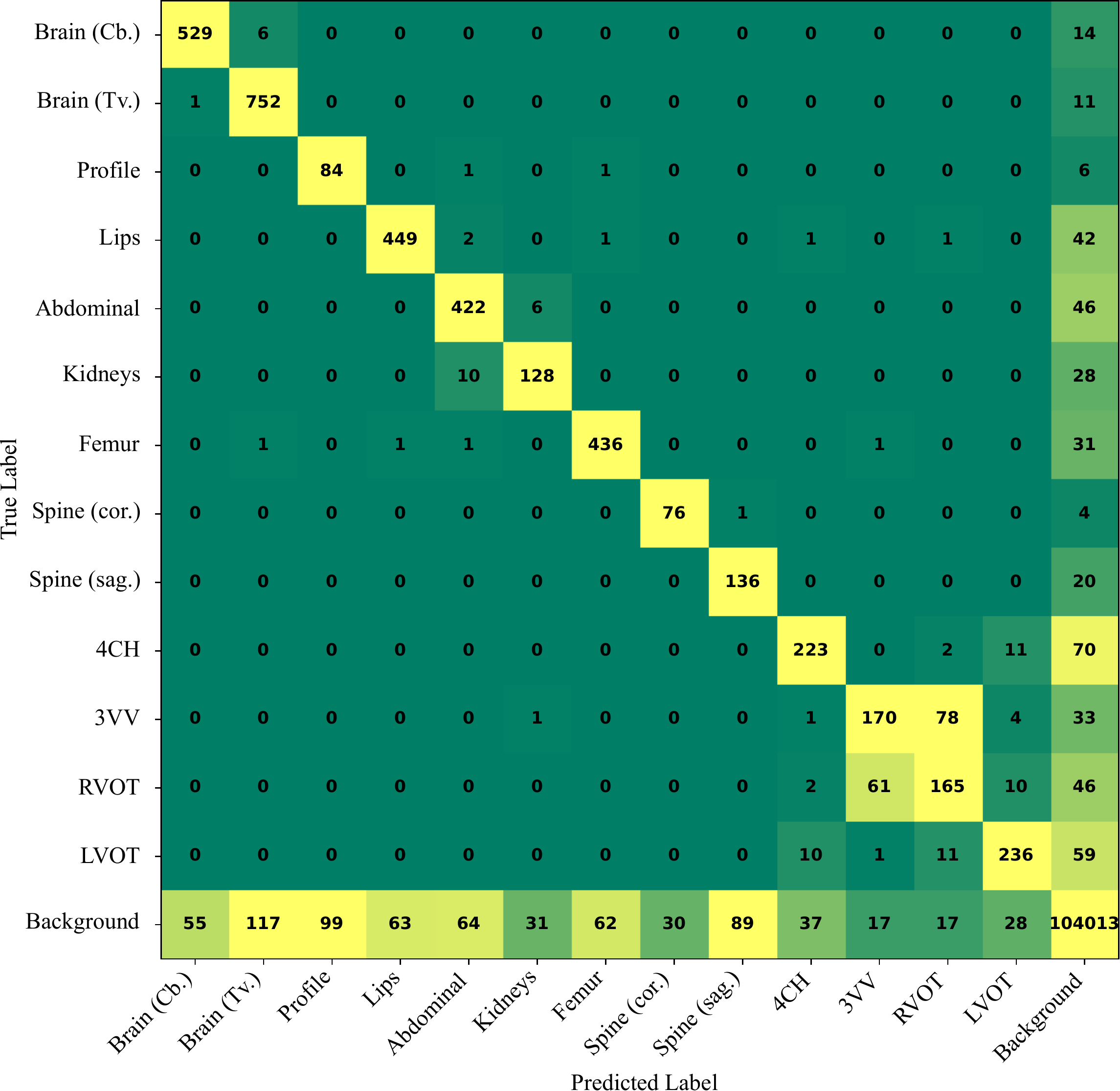}
\caption{Class confusion matrix for SonoNet-32.}
\label{fig:confusion_matrix}
\end{figure}

In Table\,\ref{tab:detection_detailed} we show the detailed classification scores for the SonoNet-32 for all the modelled categories. The right-most column lists the number of test images in each of the classes. Additionally, the class confusion matrix obtained with SonoNet-32 is shown in Fig. \ref{fig:confusion_matrix}. The results reported for this classification experiment give an indication of how the method performs in a realistic scenario. The overall ratio of standard planes to background frames is approximately 1:24 meaning that in a video on average 1 second of any of the standard views is followed by 24 seconds of background views. This is a realistic reflection of what we observe in clinical practice.

\begin{table}[t]
 \caption{Detailed classification scores for SonoNet-32}
\centering
\begin{tabular}{ l c c c c }
 \hline
 {Class} & {Precision } & {Recall } & {F1-score} & {\# Images } \\
 \hline
  Brain (Cb.)  & 0.90 & 0.96 & 0.93 & 549 \\
  Brain (Tv.)  & 0.86 & 0.98 & 0.92 & 764 \\
  Profile      & 0.46 & 0.91 & 0.61 & 92 \\
  Lips         & 0.88 & 0.91 & 0.89 & 496 \\
  Abdominal    & 0.93 & 0.90 & 0.92 & 474 \\
  Kidneys      & 0.77 & 0.77 & 0.77 & 166 \\
  Femur        & 0.87 & 0.93 & 0.90 & 471 \\
  Spine (cor.) & 0.72 & 0.94 & 0.81 & 81 \\
  Spine (sag.) & 0.60 & 0.87 & 0.71 & 156 \\
  4CH          & 0.81 & 0.73 & 0.77 & 306 \\
  3VV          & 0.68 & 0.59 & 0.63 & 287 \\
  RVOT         & 0.60 & 0.58 & 0.59 & 284 \\
  LVOT         & 0.82 & 0.74 & 0.78 & 317 \\
  Background   & 1.00 & 0.99 & 0.99 & 104722 \\ 
 \hline
 \end{tabular}
 \label{tab:detection_detailed}
\end{table}

\begin{figure}[!t]
\centering
\includegraphics[width=1.0\columnwidth]{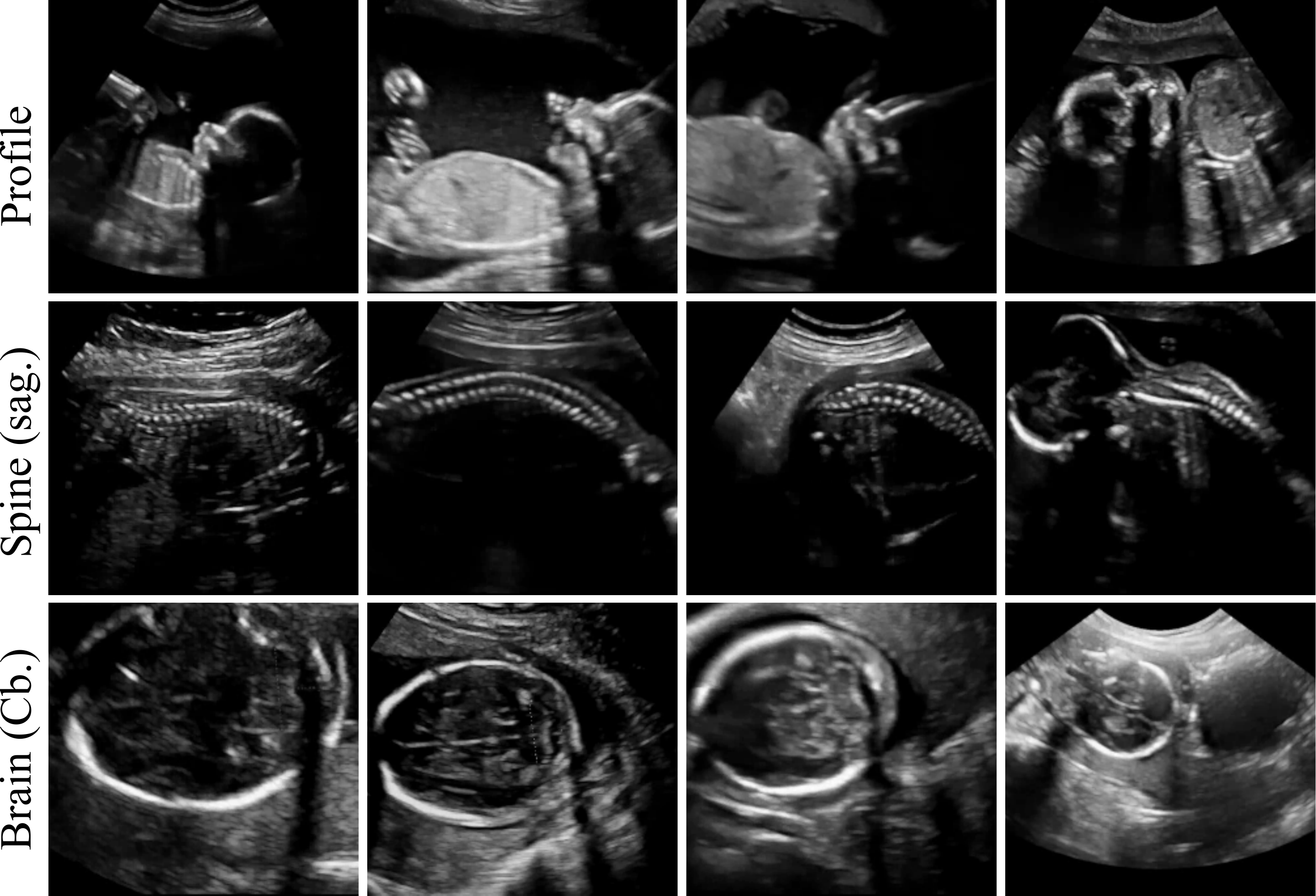}
\caption{Examples of video frames labelled as background but classified as one of three standard views. The first three columns were randomly sampled from the set of false positives and are in fact correct detections. The last column shows manually selected true failure cases.}
\label{fig:false_negatives}
\end{figure}

Some of the most important views for taking measurements and assessing the fetus' health (in particular the brain views, the abdominal view and the femur view) were detected with F1-scores of equal to or above 0.9, which are very high scores considering the difference in number of images for the background and foreground classes. The lowest detection accuracies were obtained for the profile view, the right-ventricular outflow tract (RVOT) and the three vessel view (3VV). The two cardiac views -- which are only separated from each other by a slight change in the probe angle and are very similar in appearance -- were often confused with each other by the proposed network. This can also be seen in the confusion matrix in Fig. \ref{fig:confusion_matrix}. We also noted that for some views the method produced very high recall rates with relatively low precision. The Spine (sag.) view and the profile view were particularly affected by this. We found that for a very large fraction of those false positive images, the prediction was in fact correct, but the images had an erroneous background ground-truth label. This can be explained by the fact that the spine and profile views appear very frequently in the videos without being labelled and thus many such views were inadvertently sampled in the background class generation process. Examples of cases with correct predictions but erroneous ground-truth labels for the profile and spine (sag.) classes are shown in the first three columns of Fig. \ref{fig:false_negatives}. We observed the same effect for classes which obtained higher precision scores as well. For instance, we verified that the majority of background frames classified as Brain (Cb.) are actually true detections. Examples are also shown in Fig. \ref{fig:false_negatives}. All of the images shown in the first three columns of Fig. \ref{fig:false_negatives} are similar in quality to our ground-truth data and could be used for diagnosis. Unfortunately, it is infeasible to manually verify all background images. We therefore conclude that the precision scores (and consequently F1-scores) reported in Tables \ref{tab:detection_scores_all} and \ref{tab:detection_detailed} can be considered a lower bound of the true performance.

For a qualitative evaluation, we also annotated a number of videos from our test cohort using the SonoNet-32. Two example videos demonstrating the SonoNet-32 in a real clinical exam are available at \url{https://www.youtube.com/watch?v=4V8V0jF0zFc} and \url{https://www.youtube.com/watch?v=yPCvAdOYncQ}.

\subsection{Retrospective scan plane retrieval}\label{sec:exp:detection}

We also evaluated the SonoNet-32 for retrospective retrieval of standard views on 110 random videos from the test cohort. The average duration of the recordings was 13 min 33 sec containing on average 20321 frames. The retrieved frames were manually validated by two clinical experts in obstetrics with 11 years and 3 years of experience, respectively. The time-consuming manual validation required for this experiment precluded using a larger number of videos. Table\,\ref{tab:retrieval} summarises the retrieval accuracy (TP / (P + N)) for 13 standard planes. We achieved an average retrieval accuracy of 90.09\%. As above, the most challenging views proved to be the cardiac views for which the retrieval accuracy was 82.12\%. The average accuracy for all non-cardiac views was 95.52\%. In contrast to the above experiment, the results in this section were obtained directly from full videos, and thus reflect the true performance of the method in a real scenario.

\begin{table}[t]
 \caption{Retrieval accuracy for SonoNet-32}
\centering
\begin{tabular}{ l c | l c }
 \hline
 {Class} & {Accuracy \% } & {Class } & {Accuracy \%} \\
 \hline
  Brain (Cb.)  & 96.36  & Spine (cor.)  & 95.65   \\
  Brain (Tv.)  & 100.00 & Spine (sag.)  & 96.23  \\
  Profile      & 97.73  & 4CH           & 95.00  \\
  Lips         & 92.59  & 3VV           & 81.90  \\
  Abdominal    & 88.99  & RVOT          & 73.08  \\
  Kidneys      & 78.38  & LVOT          & 78.50  \\
  Femur        & 96.70  & {\bf Average} & {\bf 90.09}  \\
 \hline
 \end{tabular}
 \label{tab:retrieval}
\end{table}

\begin{figure*}[!t] 
\centering
\includegraphics[width=1.0\textwidth]{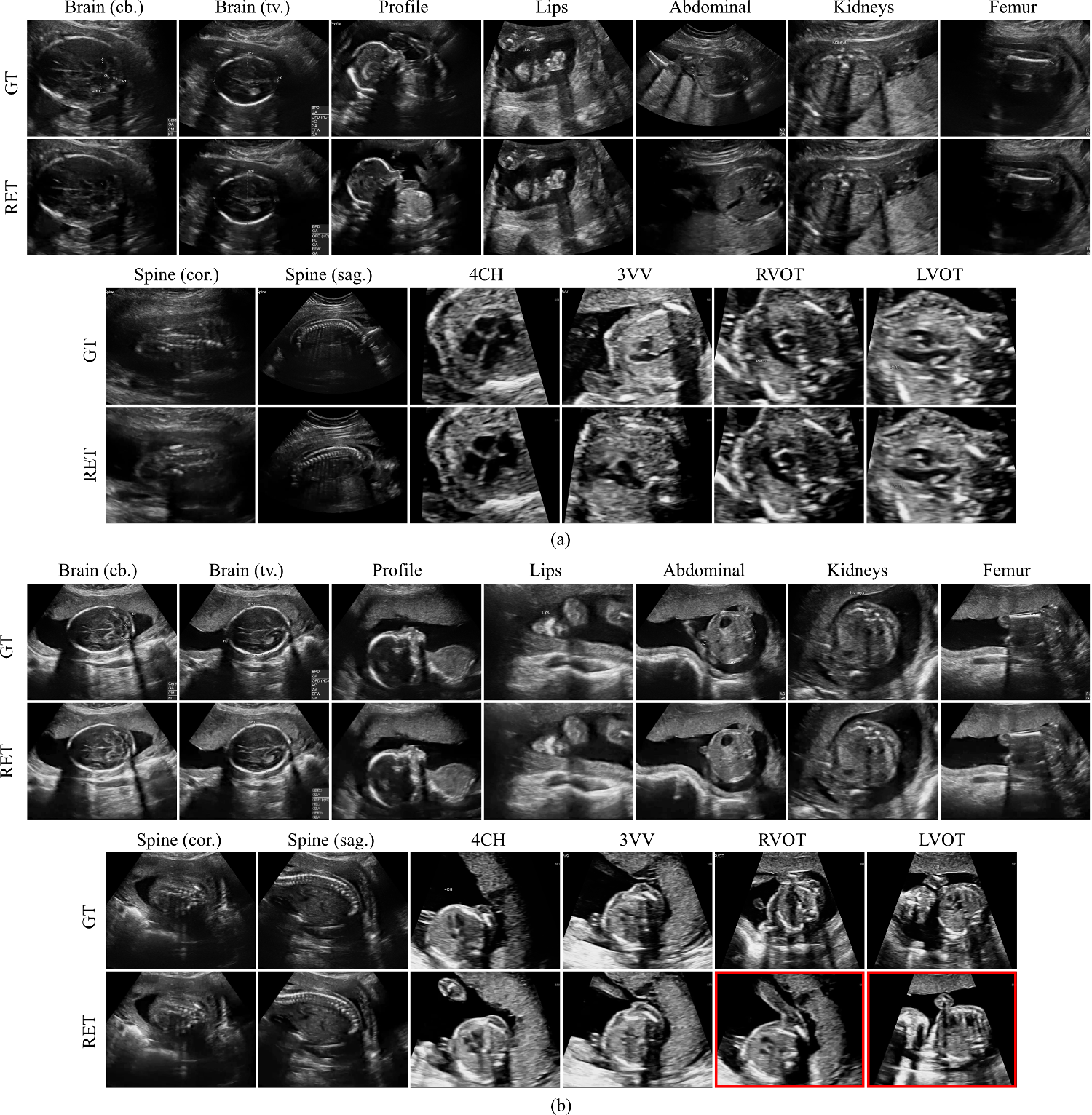} 
\caption{Results of retrospective retrieval for two example subjects. The respective top rows show the ground truth (GT) saved by the sonographer. The bottom rows show the retrieved (RET) frames. For subject (a) all frames have been correctly retrieved. For subject (b) the frames marked with red have been incorrectly retrieved.}  
\label{fig:detection_collage}
\end{figure*}  

The retrieved frames for two cases from the test cohort are shown in Fig. \ref{fig:detection_collage} along with the ground truth (GT) frames saved by the sonographers. In the case shown in Fig. \ref{fig:detection_collage}a, all views have been correctly retrieved. It can be seen that most of the retrieved frames either matched the GT exactly or were of equivalent quality. We observed this behaviour through-out the test cohort. However, a number of wrong retrievals occasionally occurred. In agreement with the quantitative results in Tab. \ref{tab:retrieval}, we noted that cardiac views were affected the most. Fig. \ref{fig:detection_collage}b shows a case for which two cardiac views have been incorrectly retrieved (marked in red). 

\subsection{Weakly supervised localisation}

We quantitatively evaluated the weakly supervised localisation using SonoNet-32 on 50 images from each of the 13 modelled standard scan planes. The 650 images were manually annotated with bounding boxes which were used as ground truth. We employed the commonly used intersection over union (IOU) metric to measure the similarity of the automatically estimated bounding box to the ground truth~\cite{everingham2010pascal}. Table\,\ref{tab:localisation} summarises the results. As in \cite{everingham2010pascal}, we counted a bounding box as correct if its IOU with the ground truth was equal to or greater than 0.5. Using this metric we found that on average 77.8\% of the automatically retrieved bounding boxes were correct. Cardiac views were the hardest to localise with an average accuracy of 62.0\%. The remaining views obtained an average localisation accuracy of 84.9\%.  

\begin{table}[t]
 \caption{Localisation evaluation: IOU and accuracy for all modelled standard views.}
\centering
\begin{tabular}{ l c c c}
 \hline
 {Class} & {Mean IOU } & {Std. IOU} & {Cor. \%} \\
 \hline
  Brain (Cb.)   & 0.73       & 0.11       & 94 \\
  Brain (Tv.)   & 0.79       & 0.11       & 96 \\
  Profile       & 0.60       & 0.13       & 78 \\
  Lips          & 0.64       & 0.22       & 78 \\
  Abdominal     & 0.68       & 0.14       & 94 \\
  Kidneys       & 0.58       & 0.15       & 74 \\
  Femur         & 0.61       & 0.16       & 76 \\
  Spine (cor.)  & 0.61       & 0.16       & 82 \\
  Spine (sag.)  & 0.69       & 0.12       & 92 \\
  4CH           & 0.53       & 0.14       & 64 \\
  3VV           & 0.54       & 0.17       & 60 \\
  RVOT          & 0.48       & 0.20       & 54 \\
  LVOT          & 0.54       & 0.16       & 70 \\
  {\bf Average} & {\bf 0.62} & {\bf 0.15} & {\bf 77.8} \\
 \hline
 \end{tabular}
 \label{tab:localisation} 
\end{table} 

\begin{figure*}[!t]
\centering
\includegraphics[width=1.0\textwidth]{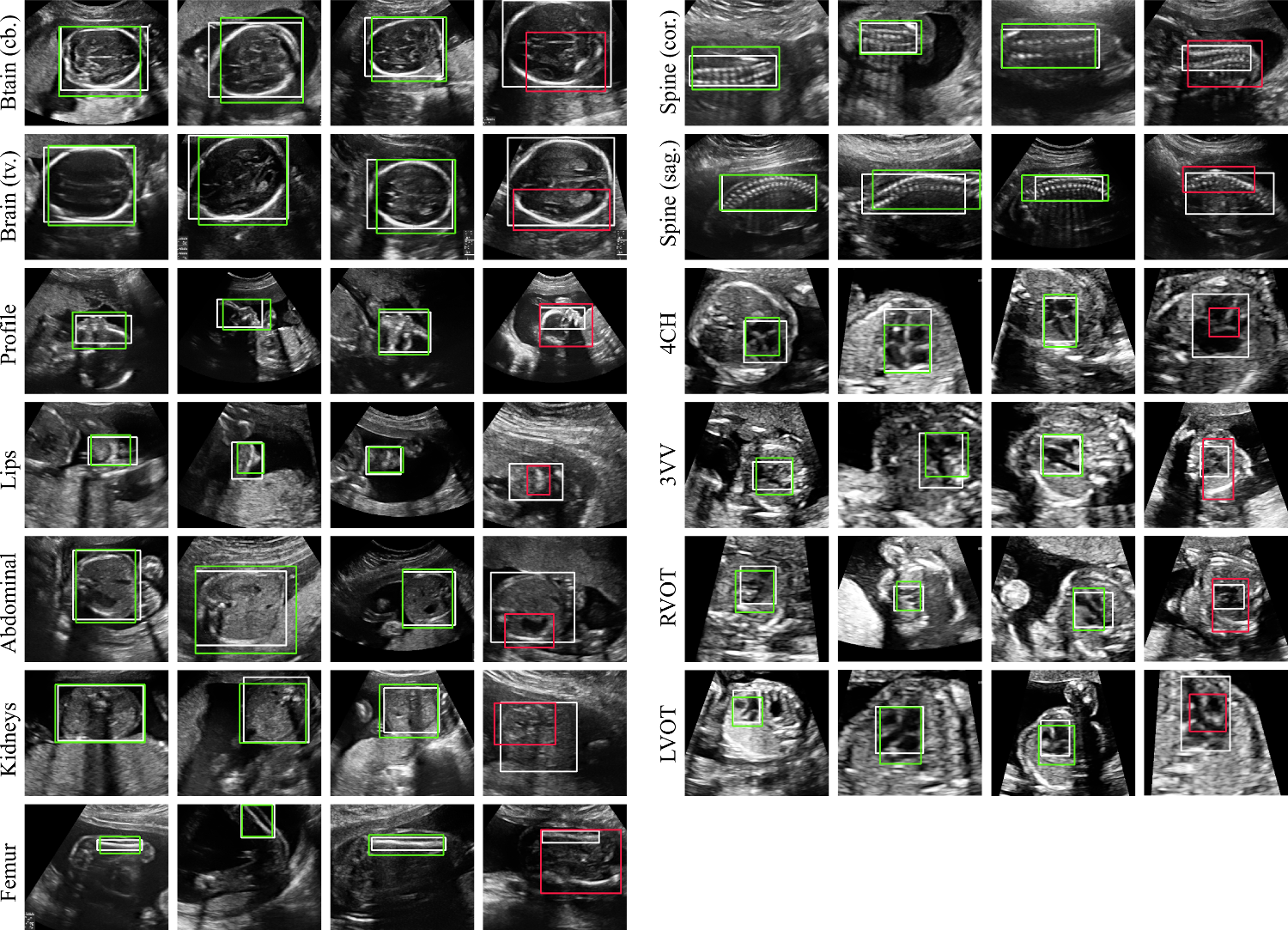}
\caption{Examples of weakly supervised localisation using the SonoNet-32. The first three columns for each view show correct bounding boxes marked in green ($IOU \geq 0.5$), the respective last columns shows an example of an incorrect localisation marked in red ($IOU < 0.5$). The ground truth bounding boxes are shown in white.}  
\label{fig:localisation_collage}  
\end{figure*}

In Fig. \ref{fig:localisation_collage} we show examples of retrieved bounding boxes for each of the classes. From these examples, it can be seen that our proposed method was able to localise standard planes
which are subject to great variability in scale and appearance. Qualitatively very good results were achieved for small structures such as the lips or the femur. The reason why this was not reflected in the quantitative results in Table\,\ref{tab:localisation} was that the IOU metric more is more sensitive to small deviations in small boxes than in large ones. 

We noted that the method was relatively robust to artefacts and performed well in cases where it may be hard for non-experts to localise the fetal anatomy. For instance, the lips view in the third column of Fig. \ref{fig:localisation_collage} and the RVOT view in the second column were both correctly localised. 

The last column for each structure in Fig. \ref{fig:localisation_collage} shows cases with incorrect ($IOU < 0.5$) localisation. It can be seen that the method almost never failed entirely to localise the view. Rather, the biggest source of error was inaccurate bounding boxes. In many cases the saliency maps were dominated by the most important feature for detecting this view, which caused the method to focus only on that feature at the expense of the remainder of the view. An example is the stomach in the abdominal view shown in the fourth column of Fig. \ref{fig:localisation_collage}. Another example is the brain (tv.) view, for which the lower parts -- where the ventricle is typically visualised -- was much more important for the detection. In other cases, regions outside of the object also appeared in the saliency map, which caused the bounding box to overestimate the extent of the fetal target structures. An example is the femur view, where the other femur also appeared in the image and caused the bounding box to cover both. 

An example video demonstrating the real-time localisation for a representative case can be viewed at \url{https://www.youtube.com/watch?v=yPCvAdOYncQ}.

\section{Discussion and Conclusion}

In this paper, we presented the first real-time framework for the detection and bounding box localisation of standard views in freehand fetal ultrasound. Notably, the localisation task can be performed without the need for bounding boxes during training. Our proposed SonoNet employs a very deep convolutional neural network, based on the widely used VGG16 architecture, but optimised for real-time performance and accurate localisation from category-specific saliency maps. 

We showed that the proposed network achieves excellent results for real-time annotation of 2D ultrasound frames and retrospective retrieval on a very challenging dataset. 

Future work will focus on including the temporal dimension in the training and prediction framework as was done for sweep data in \cite{chen2015automatic} and for fetal cardiac videos in \cite{bridge2017automated}. We expect that especially the detection of cardiac views may benefit from motion information. 

We also demonstrated the method's ability for real-time, robust localisation of the respective views in a frame. Currently, the localisation is based purely on the confidence maps shown in Fig. \ref{fig:saliency_post_processing}. Although, this already leads to very accurate localisation, we speculate that better results may be obtained by additionally taking into account the pixel intensities of the original images. Potentially, the proposed localisation method could also be combined using a multi-instance learning framework in order to incorporate the image data into the bounding box prediction~\cite{ren2016weakly}. 

We also note that the confidence maps could potentially be used in other ways, for instance, as a data term for a graphical model for semantic segmentation~\cite{boykov2001fast}. 

The pretrained weights for all of the network architectures compared in this paper are available at \url{https://github.com/baumgach/SonoNet-weights}.

\ifCLASSOPTIONcaptionsoff  
  \newpage
\fi

\bibliographystyle{plain}
\bibliography{refs}

\begin{thebibliography}{10}

\bibitem{abuhamad2008automated}
A.~Abuhamad, P.~Falkensammer, F.~Reichartseder, and Y.~Zhao.
\newblock Automated retrieval of standard diagnostic fetal cardiac ultrasound
  planes in the second trimester of pregnancy: a prospective evaluation of
  software.
\newblock {\em Ultrasound Obst Gyn}, 31(1):30--36, 2008.

\bibitem{baumgartner2016real}
C.F. Baumgartner, K.~Kamnitsas, J.~Matthew, S.~Smith, B.~Kainz, and
  D.~Rueckert.
\newblock Real-time standard scan plane detection and localisation in fetal
  ultrasound using fully convolutional neural networks.
\newblock In {\em Proc MICCAI}, pages 203--211. Springer, 2016.

\bibitem{boykov2001fast}
Y.~Boykov, O.~Veksler, and R.~Zabih.
\newblock Fast approximate energy minimization via graph cuts.
\newblock {\em IEEE T Pattern Anal}, 23(11):1222--1239, 2001.

\bibitem{bridge2017automated}
C.P. Bridge, C.~Ioannou, and J.A. Noble.
\newblock Automated annotation and quantitative description of ultrasound
  videos of the fetal heart.
\newblock {\em Med Image Anal}, 36:147--161, 2017.

\bibitem{bridge2015object}
C.P. Bridge and J.A. Noble.
\newblock Object localisation in fetal ultrasound images using invariant
  features.
\newblock In {\em Proc ISBI}, pages 156--159. IEEE, 2015.

\bibitem{chan2009volumetric}
L.W. Chan, T.Y. Fung, T.Y. Leung, D.S. Sahota, and T.K. Lau.
\newblock Volumetric {(3D)} imaging reduces inter- and intraobserver variation
  of fetal biometry measurements.
\newblock {\em Ultrasound Obst Gyn}, 33(4):447--452, 2009.

\bibitem{chen2015automatic}
H.~Chen, Q.~Dou, D.~Ni, J-Z. Cheng, J.~Qin, and P-A. Li, S.and~Heng.
\newblock Automatic fetal ultrasound standard plane detection using knowledge
  transferred recurrent neural networks.
\newblock In {\em Proc MICCAI}, pages 507--514. Springer, 2015.

\bibitem{chen2016mitosis}
H.~Chen, Q.~Dou, X.~Wang, J.~Qin, and P-A. Heng.
\newblock Mitosis detection in breast cancer histology images via deep cascaded
  networks.
\newblock In {\em Proc AAAI}, pages 1160--1166. AAAI Press, 2016.

\bibitem{chen2015standard}
H.~Chen, D.~Ni, J.~Qin, S.~Li, X.~Yang, T.~Wang, and P.A. Heng.
\newblock Standard plane localization in fetal ultrasound via domain
  transferred deep neural networks.
\newblock {\em IEEE J Biomed Health Inform}, 19(5):1627--1636, 2015.

\bibitem{el2010images}
A.~El-Zaart.
\newblock Images thresholding using isodata technique with gamma distribution.
\newblock {\em Pattern Recognit Image Anal}, 20(1):29--41, 2010.

\bibitem{everingham2010pascal}
M.~Everingham, L.~Van~Gool, C.K.I. Williams, J.~Winn, and A.~Zisserman.
\newblock The pascal visual object classes ({VOC}) challenge.
\newblock {\em Int J Comput Vision}, 88(2):303--338, 2010.

\bibitem{he2009learning}
H.~He and E.A. Garcia.
\newblock Learning from imbalanced data.
\newblock {\em IEEE T Knowl Data En}, 21(9):1263--1284, 2009.

\bibitem{he2015deep}
K.~He, X.~Zhang, S.~Ren, and J.~Sun.
\newblock Deep residual learning for image recognition.
\newblock {\em arXiv preprint arXiv:1512.03385}, 2015.

\bibitem{hill2015disparities}
G.D. Hill, J.R. Block, J.B. Tanem, and M.A. Frommelt.
\newblock Disparities in the prenatal detection of critical congenital heart
  disease.
\newblock {\em Prenatal Diag}, 35(9):859--863, 2015.

\bibitem{ioffe2015batch}
S.~Ioffe and C.~Szegedy.
\newblock Batch normalization: Accelerating deep network training by reducing
  internal covariate shift.
\newblock {\em arXiv preprint arXiv:1502.03167}, 2015.

\bibitem{kang2014fully}
K.~Kang and X.~Wang.
\newblock Fully convolutional neural networks for crowd segmentation.
\newblock {\em arXiv preprint arXiv:1411.4464}, 2014.

\bibitem{krizhevsky2012imagenet}
A.~Krizhevsky, I.~Sutskever, and G.E. Hinton.
\newblock Imagenet classification with deep convolutional neural networks.
\newblock In {\em Adv Neur In.}, pages 1097--1105, 2012.

\bibitem{lin2013network}
M.~Lin, Q.~Chen, and S.~Yan.
\newblock Network in network.
\newblock {\em arXiv:1312.4400}, 2013.

\bibitem{maraci2014searching}
M.A. Maraci, R.~Napolitano, A.~Papageorghiou, and J.A. Noble.
\newblock Searching for structures of interest in an ultrasound video sequence.
\newblock In {\em Proc MLMI}, pages 133--140. 2014.

\bibitem{maraci2015fisher}
M.A. Maraci, R.~Napolitano, A.~Papageorghiou, and J.A. Noble.
\newblock Fisher vector encoding for detecting objects of interest in
  ultrasound videos.
\newblock In {\em Proc ISBI}, pages 651--654. IEEE, 2015.

\bibitem{nhs2015FASPHandbook}
{NHS Screening Programmes}.
\newblock Fetal anomalie screen programme handbook.
\newblock 2015.

\bibitem{ni2013selective}
D.~Ni, T.~Li, X.~Yang, J.~Qin, Sh. Li, C-T. Chin, S.~Ouyang, T.~Wang, and
  S.~Chen.
\newblock Selective search and sequential detection for standard plane
  localization in ultrasound.
\newblock In {\em International MICCAI Workshop on Computational and Clinical
  Challenges in Abdominal Imaging}, pages 203--211. Springer, 2013.

\bibitem{ni2014standard}
D.~Ni, X.~Yang, X.~Chen, C-T. Chin, S.~Chen, P-A. Heng, S.~Li, J.~Qin, and
  T.~Wang.
\newblock Standard plane localization in ultrasound by radial component model
  and selective search.
\newblock {\em Ultrasound Med Biol}, 40(11):2728--2742, 2014.

\bibitem{oquab2015object}
M.~Oquab, L.~Bottou, I.~Laptev, and J.~Sivic.
\newblock Is object localization for free? - {Weakly}-supervised learning with
  convolutional neural networks.
\newblock In {\em Proc CVPR}, pages 685--694, 2015.

\bibitem{pinheiro2015image}
P.O. Pinheiro and R.~Collobert.
\newblock From image-level to pixel-level labeling with convolutional networks.
\newblock In {\em Proc CVPR}, pages 1713--1721, 2015.

\bibitem{ren2016weakly}
W.~Ren, K.~Huang, D.~Tao, and T.~Tan.
\newblock Weakly supervised large scale object localization with multiple
  instance learning and bag splitting.
\newblock {\em IEEE T Pattern Anal}, 38(2):405--416, 2016.

\bibitem{salomon2011practice}
L.J. Salomon, Z.~Alfirevic, V.~Berghella, C.~Bilardo, K-Y. Leung, G.~Malinger,
  H.~Munoz, et~al.
\newblock Practice guidelines for performance of the routine mid-trimester
  fetal ultrasound scan.
\newblock {\em Ultrasound Obst Gyn}, 37(1):116--126, 2011.

\bibitem{simonyan2013deep}
K.~Simonyan, A.~Vedaldi, and A.~Zisserman.
\newblock Deep inside convolutional networks: Visualising image classification
  models and saliency maps.
\newblock {\em arXiv preprint arXiv:1312.6034}, 2013.

\bibitem{simonyan2014very}
K.~Simonyan and A.~Zisserman.
\newblock Very deep convolutional networks for large-scale image recognition.
\newblock {\em arXiv preprint arXiv:1409.1556}, 2014.

\bibitem{sor2014sonographer}
{Society of Radiographers}.
\newblock Sonographer workforce survey analysis.
\newblock pages 28--35, 2014.

\bibitem{springenberg2014striving}
J.T. Springenberg, A.~Dosovitskiy, T.~Brox, and M.~Riedmiller.
\newblock Striving for simplicity: The all convolutional net.
\newblock {\em arXiv:1412.6806}, 2014.

\bibitem{binocar2014congenital}
{Springett, A. and Budd, J. and Draper, E.S and Kurinczuk, J.J. and Medina, J.
  and Ranking, J and Rounding, C. and Tucker, D. and Wellesley, D. and
  Wreyford, B and Morris, J.K.}
\newblock Congenital anomaly statistics 2012: {E}ngland and {W}ales.
\newblock 2014.

\bibitem{telea2004image}
A.~Telea.
\newblock An image inpainting technique based on the fast marching method.
\newblock {\em J Graph Tools}, 9(1):23--34, 2004.

\bibitem{yaqub2015guided}
M.~Yaqub, B.~Kelly, A.T. Papageorghiou, and J.A. Noble.
\newblock Guided random forests for identification of key fetal anatomy and
  image categorization in ultrasound scans.
\newblock In {\em Proc MICCAI}, pages 687--694. Springer, 2015.

\bibitem{zhang2012intelligent}
L.~Zhang, S.~Chen, C.T. Chin, T.~Wang, and S.~Li.
\newblock Intelligent scanning: Automated standard plane selection and
  biometric measurement of early gestational sac in routine ultrasound
  examination.
\newblock {\em Med Phys}, 39(8):5015--5027, 2012.

\bibitem{zhou2006training}
Z-H. Zhou and X-Y. Liu.
\newblock Training cost-sensitive neural networks with methods addressing the
  class imbalance problem.
\newblock {\em IEEE T Knowl Data En}, 18(1):63--77, 2006.

\end{thebibliography}

\end{document}